\documentclass[10pt,twocolumn,letterpaper]{article}
\pdfoutput=1

\usepackage{iccv}
\usepackage{times}
\usepackage{epsfig}
\usepackage{graphicx}
\usepackage{amsmath}
\usepackage{amssymb}
\usepackage{xcolor}
\usepackage{multirow}
\usepackage{tabularx}
\usepackage{float}
\usepackage[utf8]{inputenc}
\usepackage[T1]{fontenc}
\usepackage{diagbox}

\usepackage[pagebackref=true,breaklinks=true,letterpaper=true,colorlinks,bookmarks=false]{hyperref}

\iccvfinalcopy 

\def\iccvPaperID{21} 

\ificcvfinal\fi

\begin{document}

\title{Monocular Vision-based Prediction of Cut-in Maneuvers with LSTM Networks}

\author{Yagiz Nalcakan$^{1,2}$, Yalin Bastanlar$^{1}$\\
$^{1}$Izmir Institute of Technology\\
Urla, Izmir, Turkey\\
$^{2}$TTTech Auto Turkey Software\\
Izmir, Turkey\\
{\tt\small \{yagiznalcakan,yalinbastanlar\}@iyte.edu.tr}
}

\maketitle
\ificcvfinal\thispagestyle{empty}\fi

\begin{abstract}
   Advanced driver assistance and automated driving systems should be capable of predicting and avoiding dangerous situations. This study proposes a method to predict potentially dangerous cut-in maneuvers happening in the ego lane. We follow a computer vision-based approach that only employs a single in-vehicle RGB camera, and we classify the target vehicle's maneuver based on the recent video frames. Our algorithm consists of a CNN-based vehicle detection and tracking step and an LSTM-based maneuver classification step. It is more computationally efficient than other vision-based methods since it exploits a small number of features for the classification step rather than feeding CNNs with RGB frames. We evaluated our approach on a publicly available driving dataset and a lane change detection dataset.
    We obtained 0.9585 accuracy with side-aware two-class (cut-in vs. lane-pass) classification models. Experiment results also reveal that our approach outperforms state-of-the-art approaches when used for lane change detection.
\end{abstract}

\section{Introduction}

Prediction of intended maneuvers of surrounding vehicles is an important research area that supports the development of Advanced Driver Assistance Systems (ADAS). Also, it is one of the challenging problems on reaching fully driverless vehicles. Moreover, statistical data show that unexpected maneuvers of drivers on highways may lead to deadly accidents. According to U.S. Department of Transportation, National Highway Traffic Safety Administration's (NHTSA) 2018 report on "Driving Behaviors Reported For Drivers And Motorcycle Operators Involved In Fatal Crashes"~\cite{Report2018}, one of the top-3 reasons of fatal crashes is failure to keep the vehicle in proper lane. Therefore, early prediction of risky lane-change maneuvers of surrounding vehicles can help drivers to avoid fatal crashes on the road.

Detection and distance measurements for surrounding vehicles can be obtained via different sensors including radar, camera, and LiDAR. Each of these sensors has pros and cons compared to each other. For example, although LiDAR can detect much smaller objects and generate more detailed images compared to the others, it is still an expensive sensor. Radar has advantages on extreme illumination and weather conditions but its field of view is generally narrow and its output is noisy requiring cleaning~\cite{Sivaraman2013}. A camera, which is the sensor we use in this study, is cheap, easily accessible and it enables us to obtain a variety of information (such as color, speed, distance, depth, etc.) at the same time if accompanied with powerful computer vision techniques. 

In our study, we focus on vehicles in front and we only employ a single in-vehicle forward-looking RGB camera. This brings  simplicity to our approach compared to other studies in the literature that use camera, radar, and LiDAR sensors (\cite{Deo2018,Garcia2012,Kasper2012}). Since there is no benchmark dataset for the classification of potentially dangerous cut-in maneuvers in traffic, we have prepared a classification dataset with the videos of the publicly available Berkeley Deep Drive dataset~\cite{bdd100k}, which consists of videos that are collected via the front camera of the vehicles on highways of various cities. 
We have cut and labeled 875 video clips containing vehicle maneuvers belonging to cut-in or lane-pass classes. These video clips cover two seconds of action. In our experiments, we represented this duration with varying number of frames (15, 30, 45 or 60). As the number of frames increases, it becomes a dense representation but also it requires more computation. We made labeled clips and source code of our methods publicly available\footnote{https://github.com/ynalcakan/cut-in-maneuver-prediction}.
Apart from the dataset we prepared, we evaluated our method on a cut-in and lane-pass maneuver subset extracted from a lane change detection benchmark dataset~\cite{prevention_dataset}.

\begin{figure}[!htb]
      \centering
      \includegraphics[scale=0.24]{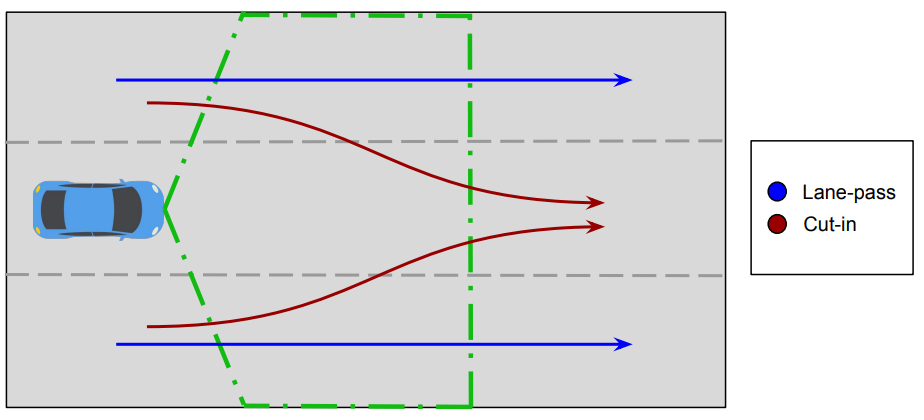}
      \caption{Lane-pass and cut-in maneuvers (green area indicates considered safety field for ego vehicle.)}
      \label{fig:cutin}
\end{figure}

We classify the maneuver of the vehicles in front whether they are cutting-in to ego-vehicle's lane or keeping their own lane (Figure \ref{fig:cutin}). In our experiments, we evaluated several models for two-class classification of cut-in and lane-pass maneuvers, as well as 3-class models to discriminate left-hand side and right-hand side cut-ins. There is no doubt that the proposition of classifying maneuvers into two or three is an oversimplification of the real life cases. However, this scheme is enough from the viewpoint of predicting risky cut-in maneuvers and it enables us to compare our results with the state-of-the-art lane-change detection methods (\cite{Biparva2021, Llorca2020, Izquierdo2019}).

\begin{figure}[h!]
      \centering
      \includegraphics[scale=0.22]{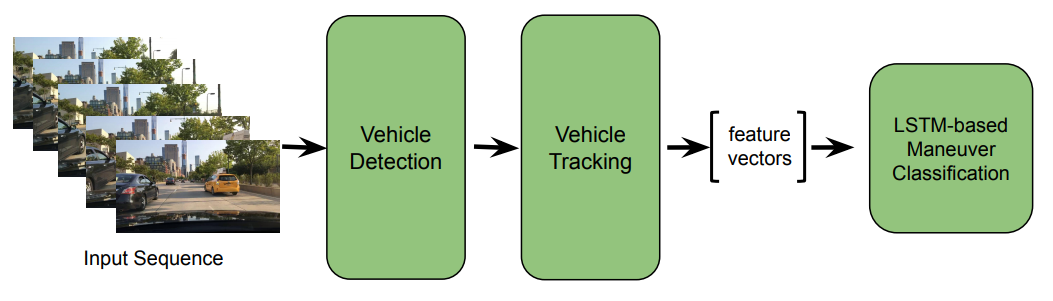}
      \caption{Overview of the proposed approach}
      \label{fig:outline}
\end{figure}

Our approach consists of three steps (Figure \ref{fig:outline}). The first step is a CNN-based vehicle detection, where we employ YOLOv4~\cite{Bochkovskiy2020} to detect vehicles in each frame of the sequence. The second step is tracking the detected vehicles using DeepSort~\cite{Wojke2017}. In the third step, extracted features from the detected and tracked bounding boxes of vehicles in front are fed into an LSTM network to be classified.

Expensive sensor setups or complex processing pipelines limit the availability and robustness of previous methods.
Thus, we state the contributions of our work as:
\begin{enumerate}
    \item Our approach employs very simple equipment (a standard field-of-view RGB camera) and does not require a  calibration procedure or any other specific adjustment depending on the ego-vehicle.
    \item Our approach is computationally cheap compared to previous work that feed CNNs with video frames and use complex architectures
    \cite{Biparva2021, Llorca2020, Izquierdo2019, Yurtsever2019}.
    Instead, we exploit a small number of features extracted from input sequence and feed an LSTM with those. We exceed the classification accuracies reported in the compared studies. Moreover, we are able to produce a classification decision every two seconds.
    
\end{enumerate}

The remainder of this paper is structured as follows: the related works are reviewed in Section 2. Section 3 has detailed information about our method on maneuver prediction and Section 4 provides information about datasets. Finally, experimental results discussed in Section 5 which is followed by the conclusion in Section 6.

\section{Related Work}

In the literature, different classifications have been made on the modeling of vehicle motions. LeF\'evre \etal~\cite{lefevre2014}  divided motion modeling approaches into three as physics-based models, maneuver-based models, and interaction-aware models by their targets' motion. As our approach considers target vehicle's maneuvers to classify its motion, it is among the maneuver-based models.

Most of the work done in the field of maneuver classification is on lane change prediction. They try to classify right lane change, left lane change and no lane change maneuvers. There are few studies on detecting risky cut-in maneuvers, which we focus on.

We would like to discuss the recent studies in the literature by dividing them into two according to their methodologies, as trajectory-based and vision-based maneuver classifications. Basically, trajectory-based maneuver classification is about projecting the trajectory of each surrounding vehicle on the ground plane using on-vehicle camera, radar or external sensors (i.e. surveillance cameras) and classifying the maneuver with this trajectory information. Vision-based classification is about extracting features from an image sequence and classifying the maneuver using these features.

\subsection{Trajectory-based Maneuver Classification}

In a study conducted before the deep learning era, Kasper \etal ~\cite{Kasper2012} proposed to model driving maneuvers using Object-oriented Bayesian Networks (OOBN). According to their experiments, the combined use of lane-related coordinate features and occupancy grids are very effective to classify driving maneuvers.
Deo \etal ~\cite{Deo2018} proposed an approach to classify maneuver trajectories by using hidden markov model (HMM) and variational Gaussian mixture model (VGMM). First, they extracted different maneuvers like lane-pass, overtake, cut-in, and drift-into-ego-lane from highway recorded videos, radar and LiDAR data. Then they classified all trajectories using VGMM. Their method reached 0.842 accuracy on all maneuvers 
and 0.559 accuracy on overtake and cut-in maneuvers. 
Scheel \etal ~\cite{Scheel2019} used the trajectories of the right lane change, left lane change, and follow maneuvers as input to an attention-based LSTM network, and they reported accuracy of prediction of each maneuver separately as 0.784 for left lane change, 0.962 for follow, and 0.679 for right lane change. Altché \etal ~\cite{Altche2018} proposed an LSTM-based method to predict future vehicle trajectories on NGSIM dataset~\cite{ngsim} in which two layer LSTM achieved better RMSE results compared to other similar approaches in two and three seconds prediction horizons.

\subsection{Vision-based Maneuver Classification}

With the increasing popularity of deep learning methods in vision, recent studies of vision-based maneuver classification generally use convolutional neural networks (CNNs) to get visual information regarding the scene. Usual practice is using a CNN as a feature extractor by feeding video frames into CNN and using an RNN or an LSTM as a classifier. In ~\cite{Izquierdo2019}, features are extracted by a CNN on region-of-interest (ROI) and width, height, center coordinate values are added to the feature vector. Then, classification of the lane change maneuvers is performed by an LSTM. Their best model achieved 0.745 accuracy. Another approach in the same study~\cite{Izquierdo2019} was converting movements of objects into contours in an RGB image and feeding CNNs with this motion history image. However, performance was worse. 

Another study that first crops ROIs from the original frames~\cite{Llorca2020} exploited two modes of input video, which are high frame rate video itself and its optical flows. They compared two-stream CNNs and spatio-temporal multiplier networks. In a follow-up study~\cite{Biparva2021}, authors also included slow-fast network (the one uses videos of high and low frame rate) into the comparison which achieved 0.908 accuracy and performed slightly better than other alternatives. 


The studies discussed above have used the Prevention dataset~\cite{prevention_dataset} to test their methods. In other studies that perform vision-based maneuver classification on different datasets, Lee \etal ~\cite{Lee2017} proposed a method that can be used for adaptive cruise control, which uses a front-faced radar and camera outputs to infer the lane change maneuvers. In their method, they convert the traffic scenes to a simplified bird's eye view (SBV) and those SBVs are given as input to a CNN network to predict lane keeping, right cut-in and left cut-in intentions. Yurtsever \etal ~\cite{Yurtsever2019} proposed a deep learning-based action recognition framework for classifying dangerous lane change behavior in video captured by an in-car camera. They used a pre-trained Mask R-CNN model to segment vehicles in the scene and a CNN+LSTM model to classify the behaviors as dangerous or safe. 

In this study, we extracted bounding boxes of surrounding vehicles with well-performing computer vision techniques. This step exists in previous vision based methods as well.
However, unlike previous methods (\cite{Biparva2021,Llorca2020,Izquierdo2019,Lee2017,Yurtsever2019}) we do not feed a feature extractor CNN with RGB frames of the sequence. Instead, we feed an LSTM with a feature vector consisting of bounding box data, which is very fast. With a 15-frame model we can make an estimate for every two seconds.

To the best of our knowledge, there is no vision-based cut-in/lane-pass classification accuracy reported in the literature but we can compare our method's performance with previously reported lane-change classification methods (that does not differ if a vehicle is entering or existing ego-lane). Thus, in addition to the cut-in detection performances, we share our method's performance on lane change prediction task on the publicly available Prevention dataset~\cite{prevention_dataset} (cf. Section 5.3).
Results indicate that with the proposed LSTM approach, a 3-class (left lane-change, right lane-change, no lane-change) classification accuracy of over 94\% can be obtained whereas previously reported lane-change classification accuracies (\cite{Biparva2021,Llorca2020,Izquierdo2019}) do not reach 92\% on the same dataset even with complex models.


\section{Methodology}

\subsection{Vehicle Detection}

\begin{figure*}[!htb]
      \centering
      \includegraphics[scale=0.54]{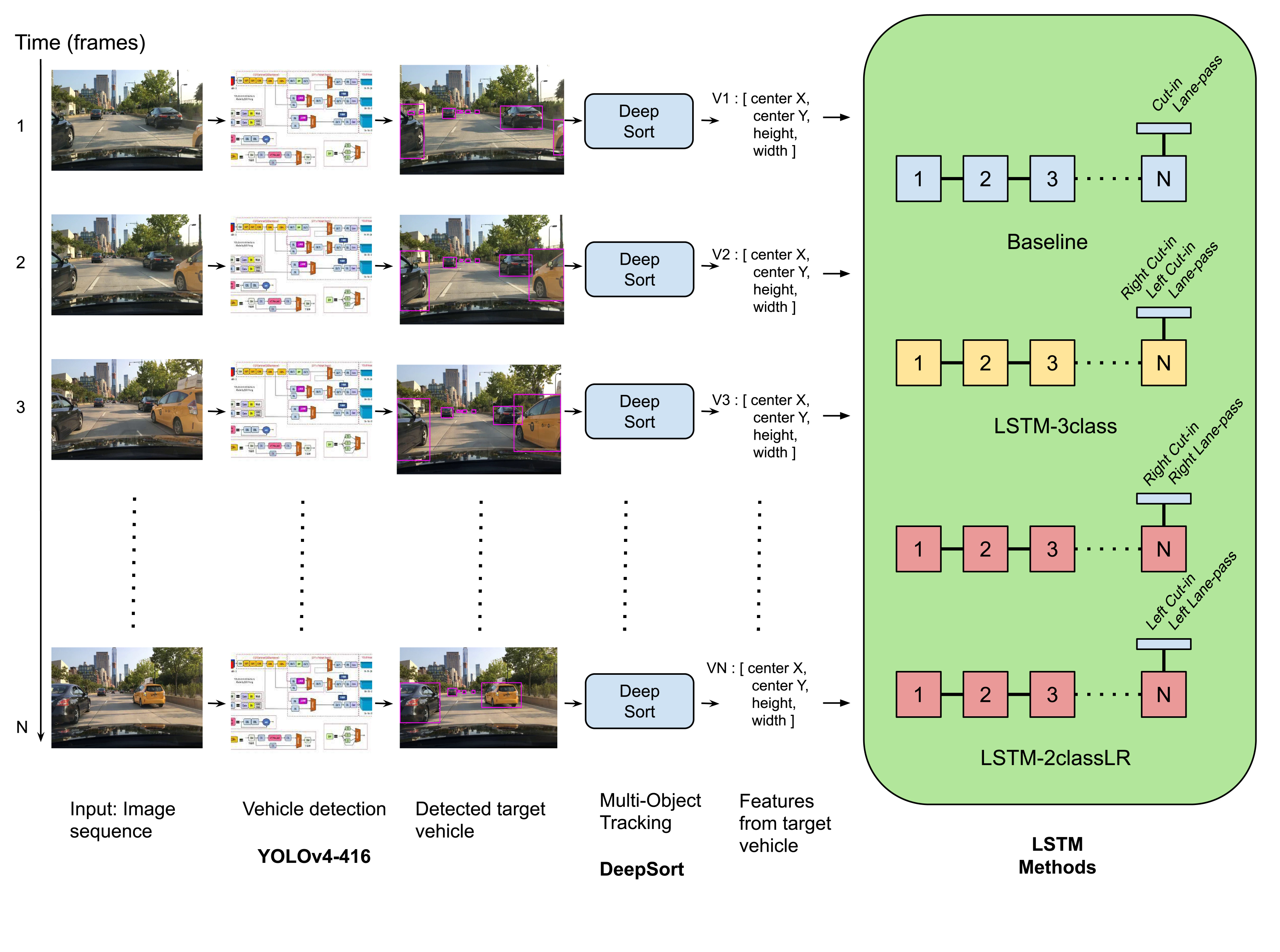}
      \caption{Pipeline of the proposed approach. Following the steps for extracting the bounding boxes of target vehicles (TVs), baseline LSTM method uses feature vectors of TVs and classifies maneuvers as cut-in or lane-pass. LSTM-3class method  classifies into 3: right cut-in, left cut-in or lane-pass classes. As a third alternative, LSTM-2classLR has two separate LSTMs for left-hand side and right-hand side TVs. We conducted experiments with varying sequence lengths (15, 30, 45 and 60 frames) all representing two seconds of the video.}
      \label{fig:pipeline}
\end{figure*}

As we extracted our features using the bounding box of the target vehicle, it was very important to collect the information regarding the surrounding vehicles with high accuracy. Therefore, as vehicle detection network, we used YOLOv4 ~\cite{Bochkovskiy2020} which is widely used in similar methods. Each frame of the image sequence is fed into the pre-trained YOLOv4 and  bounding boxes of the target vehicles are sent to the vehicle tracking step together with the same input image (Figure \ref{fig:pipeline}).

\subsection{Vehicle Tracking}

Detected vehicles per frame (bounding boxes) can have position errors caused by YOLOv4. To minimize those errors, we added a vehicle tracking step to the pipeline. Tracking is done via DeepSort ~\cite{Wojke2017} which is known for its ease of implementation with YOLO. Corrected bounding boxes of target vehicle are given as input to the feature extraction step.

\subsection{Feature Extraction and Network Architecture}

Many visual cues may allow us to predict whether any of the surrounding vehicles makes a cut-in maneuver or not. In our work, we obtain cues from the detected bounding boxes of surrounding vehicles. Specifically, we extract the center ($x$,$y$) coordinates, width and height values of the bounding box. Collected features are given to a single-layer LSTM to obtain the classification result. We tried four different sequence lengths (15, 30, 45, 60). Shorter sequence length means more sparse representation (3 out of 4 frames are neglected in 15-frame sequences) of the action but faster maneuver prediction due to decreased processing time.

As hyperparameters of the LSTM, various hidden unit sizes, batch sizes, activation unit types etc. are evaluated to find the best performing LSTM architecture. Evaluated hyperparameters are given in Table~\ref{table:hyperparameters} and proposed framework can be seen in Figure~\ref{fig:pipeline}.

\setlength{\tabcolsep}{3pt}
\setlength{\extrarowheight}{.10em}
\begin{table}[!htb]
\centering
\caption{Evaluated LSTM hyperparameters}
\label{table:hyperparameters}
\begin{tabular}{|l|l|l|l|l|} 
\hline
\begin{tabular}[c]{@{}c@{}}Hidden\\ Units\end{tabular} & \begin{tabular}[c]{@{}c@{}}Batch\\ Sizes\end{tabular} & Optimizer & Activation & Dropout \\ 
\hline
\multicolumn{1}{|c|}{\begin{tabular}[c]{@{}c@{}}60\\128\\256\\512\end{tabular}} & \multicolumn{1}{c|}{\begin{tabular}[c]{@{}c@{}}5\\10\\50\\100\end{tabular}} & \multicolumn{1}{c|}{\begin{tabular}[c]{@{}c@{}}Adam\\~RMSProp\\AdaDelta\end{tabular}} & \multicolumn{1}{c|}{\begin{tabular}[c]{@{}c@{}}ReLU\\Sigmoid\\Tanh\end{tabular}} &
\multicolumn{1}{c|}{\begin{tabular}[c]{@{}c@{}}0\\0.25\\0.5\end{tabular}} \\
\hline
\end{tabular}
\end{table}


\section{Dataset}

\subsection{BDD-100K Dataset and Labeling}

A subset of the Berkeley Deep Drive Dataset ~\cite{bdd100k} was used in this study. This dataset consists of 100K driving video that labeled for 10 different tasks (road object detection, instance segmentation, drive-able area, etc.). The videos were collected  from the front camera of the vehicles at various times of the day in New York, Berkeley, San Francisco, and Tel Aviv. We focused on lane change actions that happened at highways. 875 video sequences containing vehicle maneuvers belonging to cut-in and lane-pass classes (Figure \ref{fig:cutin}) were cut from approximately 20K videos.
The final distribution contains 405 cut-in and 470 lane-pass samples which are divided into train-validation-test datasets using 60\%-20\%-20\% split ratio. 

\begin{figure}[!htb]
      \centering
      \includegraphics[scale=0.185]{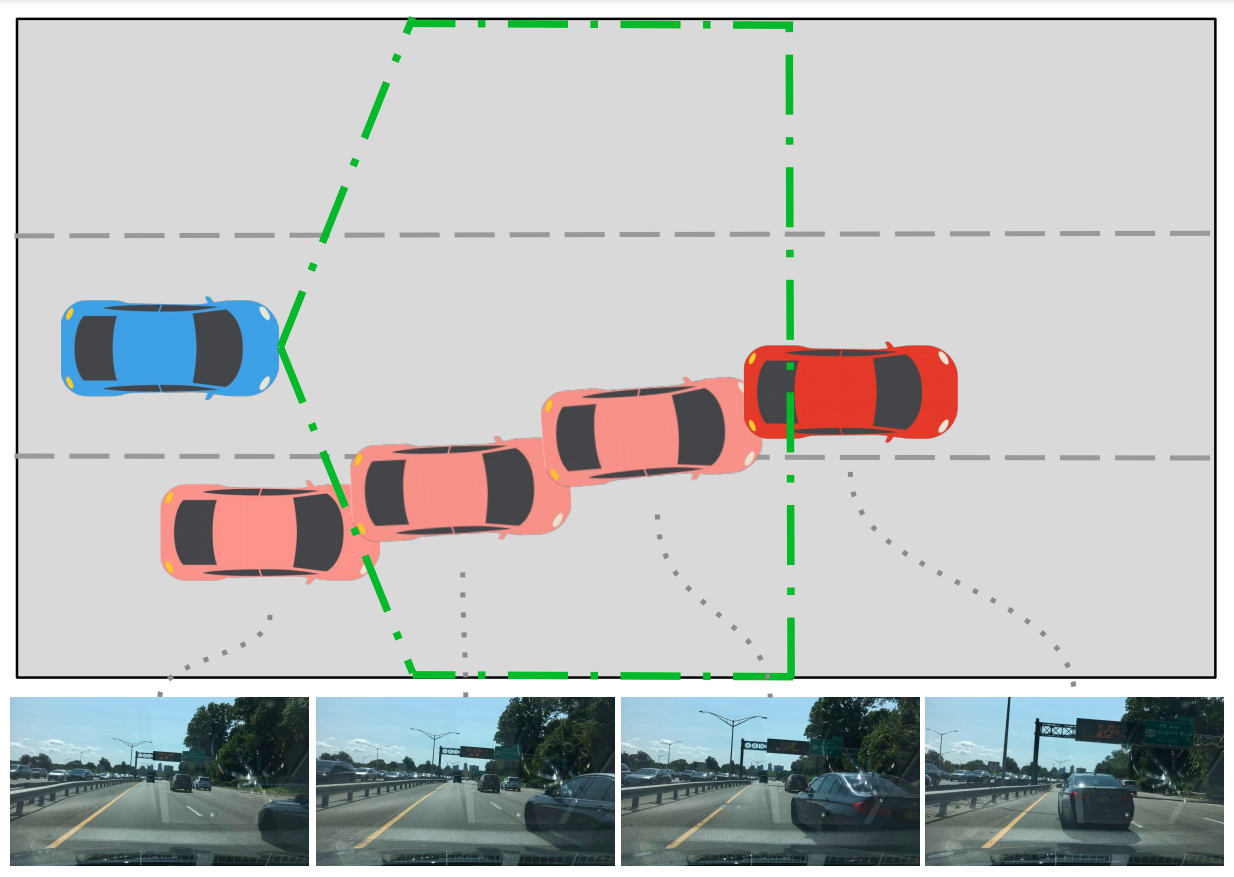}
      \caption{Start and end points of maneuvers that are labeled as cut-in }
      \label{fig:cutinlabeling}
\end{figure}

Figure \ref{fig:cutinlabeling} shows the principle while labeling cut-in samples in our dataset.
At the starting frame of the sequence, the target vehicle is on the other lane and there is no indication whether a cut-in will occur or not. 
The lane change event occurs as the target vehicle enters the safety field (the polygon that is indicated with green lines in Figure \ref{fig:cutinlabeling}).
The sequence is cut when the vehicle is entered to the ego-lane with its full body (no need to be aligned in the center). For the lane-pass class, the vehicles those pass by the ego vehicle from right-hand side or left-hand side labeled during their stay in safety field.

\subsection{Prevention Dataset and Labeling}

In 2019, Izquiredo \etal~\cite{prevention_dataset} from University of Alac\'a published a benchmark dataset called "The PREVENTION dataset: a novel benchmark for PREdiction of VEhicles iNTentIONs" for lane change detection problem. The dataset has 356 hours of driving video which is recorded in mostly highways. They provide detections, trajectories and labels additionally to the raw data. They continuously improve the dataset but the current version has only three labels for vehicle maneuvers which are "left lane change", "right lane change" and "no lane change". According to~\cite{prevention_dataset}, they are planning to add left cut-in, right cut-in, cut-out and hazardous classes to the dataset.

Since recent studies have used the Prevention Dataset \cite{Biparva2021,Llorca2020,Izquierdo2019}, we included it in our evaluation as well. We generated two separate subsets, one is a cut-in/lane-pass subset and the other is lane change detection subset. 
Number of samples used for lane change prediction and cut-in prediction tasks are shown in Table~\ref{table:distribution}. Since the number of samples of no lane change and lane change categories are unbalanced in the original dataset, the same skewness occurs in the prepared subsets as well.

\setlength{\tabcolsep}{3pt}
\setlength{\extrarowheight}{.20em}
\begin{table}[h!]
\centering
\caption{Distribution of samples extracted from the Prevention Dataset both for lane change prediction task and  cut-in/lane-pass classification task}
\vspace{1mm}
\label{table:distribution}
\begin{tabular}{|c|c|c|}
\hline
\diagbox{Class}{Task} & \begin{tabular}[c]{@{}c@{}}Lane Change\\ Prediction\end{tabular} & \begin{tabular}[c]{@{}c@{}}Cut-in/Lane-pass\\ Prediction\end{tabular} \\ \hline
No Lane Change    & 4827 & - \\
Left Lane Change  & 267 & -  \\
Right Lane Change & 200 & -  \\
Lane-pass         & -   & 1132 \\
Cut-in            & -   & 114  \\ \hline
\end{tabular}
\end{table}
\setlength{\tabcolsep}{1.4pt}

Cut-in/lane-pass subset sequences have at least 60 frames to be able to compare with our BDD-100K cut-in/lane-pass dataset. Sequences were labeled in the manner shown in Figure~\ref{fig:cutinlabeling}. For the lane change detection subset, to keep as many samples as possible, we have included sequences that have at least 50 frames. 

For train-validation-test split same split ratio of 60\%-20\%-20\% as in BDD-100K subset used. In addition, hyperparameter selection was made by selecting from the same hyperparameter pool(Table~\ref{table:hyperparameters}) with the grid search method as in the BDD-100K subset.


\section{Experimental Results}

\subsection{Cut-in$/$Lane-pass Classification Results on BDD-100K Dataset}

We evaluated our approach with different methods where classification strategy varies. As a baseline method, a single layer LSTM model is trained with four features (center ($x$,$y$) coordinates, width and height of the target vehicle's bounding box) for a side-agnostic 2-class classification, i.e. each sample is a cut-in or a lane-pass. In LSTM-3class method, samples are classified samples as left-hand side cut-in, right-hand side cut-in and lane-pass. This second strategy is closer to several lane-change prediction studies in the literature, where maneuvers were classified as left lane-change, right lane-change and no lane-change. By examining the target vehicle's center coordinates, it is straightforward to extract if it is on the left or on the right of the ego vehicle. Thus, as a third method (LSTM-2classLR), we train two networks one responsible for left-hand side maneuvers and the other for right-hand side, each performs a 2-class classification (cut-in/lane-pass).
Test performances of these 3 methods are presented in
Table~\ref{table:comparison}.

Since most previous studies reported accuracy, precision, and recall, our evaluation is based on these metrics.
For all models mentioned above, hyperparameters (Table~\ref{table:hyperparameters}) were optimized by grid search and the model with the highest accuracy on the validation set was evaluated on the test set.

Our baseline method, which classifies the sequences without separating if the maneuver is on the right or on the left, reached an accuracy of 0.9256 with 30-frame and 60-frame sequences and slightly lower accuracy for other sequence lengths. Taking into account the side of the cut-in maneuver (3 classes: right cut-in, left cut-in, lane-pass) caused a very slight decrease in the performance, achieving 0.9324 accuracy. However, when we train two separate networks for the right-hand side and left-hand side (LSTM-2classLR), the classification accuracy increased to 0.9585. The best values were obtained with 30-frame sequences, but for almost all other lengths accuracies were increased compared to the baseline and LSTM-3class methods.

\begin{table*}[h!]
\centering
\caption{Cut-in$/$Lane-pass classification results of different methods for varying sequence lengths on BDD-100K dataset}
\vspace{1mm}
\label{table:comparison}
\begin{tabular}{|c|c|c|c|c|c|c|} 
\hline
Method                         & \begin{tabular}[c]{@{}c@{}}Sequence\\ Length\end{tabular} & Accuracy        & \begin{tabular}[c]{@{}c@{}}Precision\\ (Cut-in)\end{tabular} & \begin{tabular}[c]{@{}c@{}}Recall\\ (Cut-in)\end{tabular} & \begin{tabular}[c]{@{}c@{}}Precision\\ (Lane-pass)\end{tabular} & \begin{tabular}[c]{@{}c@{}}Recall\\ (Lane-pass)\end{tabular} \\ 
\hline
\multirow{4}{*}{Baseline} & 15 & 0.8851 & 0.8551 & 0.8939 & 0.9114 & 0.8780 \\
& 30 & \textbf{0.9256} & 0.8841 & \textbf{0.9531} & \textbf{0.9620} & 0.9048 \\
& 45 & 0.9189 & \textbf{0.9275} & 0.9014 & 0.9114 & 0.9351 \\
& 60 & \textbf{0.9256} & \textbf{0.9275} & 0.9143 & 0.9241 & \textbf{0.9359} \\ 
\hline
\multirow{4}{*}{LSTM-3Class} & 15 & \textbf{0.9324} & \textbf{0.8960} & 0.9230 & \textbf{0.9620} & \textbf{0.9383} \\
& 30 & 0.9121 & 0.8667 & 0.9077 & 0.9494 & 0.9146 \\
& 45 & \textbf{0.9324} & 0.8769 & \textbf{0.9245} & 0.9494 & 0.9146 \\
& 60 & 0.9256 & 0.8829 & 0.9077 & \textbf{0.9620} & \textbf{0.9383} \\ 
\hline
\multirow{4}{*}{LSTM-2ClassLR} 
& 15 & 0.9311 & 0.9321 & 0.9021 & 0.9302 & 0.9524 \\
& 30 & \textbf{0.9585} & \textbf{0.9494} & 0.9500 & 0.9651 & \textbf{0.9648} \\
& 45 & 0.9452 & 0.9171 & 0.9483 & 0.9651 & 0.9434 \\
& 60 & 0.9519 & 0.9160 & \textbf{0.9648} & \textbf{0.9767} & 0.9439 \\
\hline
\end{tabular}
\end{table*}

\subsection{Cut-in$/$Lane-pass Classification Results on Prevention Dataset}

Each method tested on the BDD-100K dataset is also evaluated on the Prevention cut-in/lane-pass subset (cf. Section 4.2). Accuracy, precision, and recall results with different sequence lengths are shown in Table~\ref{table:comparison-pr-2class}. While all three models are highly succesful, LSTM-2ClassLR is slightly better than the others, which is consistent with the results with BDD-100K dataset. These results indicate that the success of the proposed approach is not specific to a dataset and works well on a benchmark dataset as well.

We observe occasional drops in precision and recall values of cut-in compared to those of lane-pass class. This is due to the skewness in the dataset. Since the number of lane-pass samples is much higher, model is inclined to prefer lane-pass more. In the lane-change study with this dataset \cite{Biparva2021} as well, reported lane-change precision and recall are much lower than those of no-lane-change class.

\begin{table*}[h!]
\centering
\caption{Cut-in/Lane-pass classification results of different methods for varying sequence lengths on Prevention dataset}
\vspace{1mm}
\label{table:comparison-pr-2class}
\begin{tabular}{|c|c|c|c|c|c|c|}
\hline
Method & \begin{tabular}[c]{@{}c@{}}Sequence\\ Length\end{tabular} & Accuracy & \begin{tabular}[c]{@{}c@{}}Precision\\ (Cut-in)\end{tabular}  & \begin{tabular}[c]{@{}c@{}}Recall\\ (Cut-in)\end{tabular} & \begin{tabular}[c]{@{}c@{}}Precision\\ (Lane-pass)\end{tabular}  & \begin{tabular}[c]{@{}c@{}}Recall\\ (Lane-pass)\end{tabular} \\ \hline
\multirow{4}{*}{Baseline} 
& 15 & 0.9718 & 0.9259 & 0.8333 & 0.9775 & 0.9909 \\
& 30 & \textbf{0.9799} & \textbf{0.9630} & \textbf{0.8667} & \textbf{0.9820} & \textbf{0.9954} \\
& 45 & 0.9638 & 0.8519 & 0.8214 & 0.9775 & 0.9819 \\
& 60 & 0.9759 & 0.8889 & 0.8276 & 0.9777 & 0.9865 \\ \hline
\multirow{4}{*}{LSTM-3Class} 
& 15 & 0.9638 & 0.6786 & \textbf{0.9286} & \textbf{0.9865} & 0.9733 \\
& 30 & \textbf{0.9719} & \textbf{0.9524} & 0.8958 & 0.9775 & \textbf{0.9909} \\
& 45 & 0.9558 & 0.7976 & 0.7976 & 0.9730 & 0.9730 \\
& 60 & 0.9598 & 0.6786 & 0.7500 & 0.9820 & 0.9732 \\ \hline
\multirow{4}{*}{LSTM-2ClassLR} 
& 15 & 0.9788 & 0.9285 & \textbf{0.8903} & \textbf{0.9845} & 0.9920 \\
& 30 & 0.9740 & 0.9286 & 0.8452 & 0.9791 & 0.9919 \\
& 45 & 0.9665 & 0.9286 & 0.7917 & 0.9710 & 0.9918 \\
& 60 & \textbf{0.9789} & \textbf{0.9524} & 0.8690 & 0.9818 & \textbf{0.9946} \\ \hline
\end{tabular}
\end{table*}

\subsection{Lane Change Prediction Results on Prevention Dataset}

To compare with the studies in the literature, we trained our LSTM-3Class model for lane change detection using the Prevention Dataset which ~\cite{Biparva2021}, ~\cite{Llorca2020}, and ~\cite{Izquierdo2019} used in their studies. As we explained above, LSTM-3Class model runs with 15, 30, 45 and 60-frame inputs. Since we allowed 50-frame sequences in the lane change detection subset (to keep more samples), we compare only the results of 15-frame, 30-frame, and 45-frame LSTM-3Class models with other studies. That comparison can be seen in Table~\ref{table:comparison-pr-3class}.

Even though the focus of our study is cut-in prediction, we see that if the proposed LSTM-based approach is trained for lane change prediction, its performance exceeds the previously reported performances which are 3-class lane-change prediction accuracies of 0.7440 in~\cite{Izquierdo2019}, 0.9190 in~\cite{Biparva2021}, and 0.9194 in~\cite{Llorca2020}.

\setlength{\tabcolsep}{3pt}
\setlength{\extrarowheight}{.20em}
\begin{table*}[h!]
\centering
\caption{Comparison with the previous lane change maneuver classification studies}
\vspace{1mm}
\label{table:comparison-pr-3class}
\begin{tabular}{|c|c|c|}
\hline
 & Method & Accuracy \\ \hline
\multirow{1}{*}{Biparva \etal~\cite{Biparva2021}} 
& Spatiotemporal Multiplier Network & 0.9190 \\ \hline
\multirow{1}{*}{Izquierdo \etal~\cite{Izquierdo2019}} 
& GoogleNet + LSTM & 0.7440 \\ \hline
\multirow{1}{*}{Fernandez-Llorca \etal~\cite{Llorca2020}} 
& Spatiotemporal Multiplier Network & 0.9194 \\ \hline
\multirow{1}{*}{Ours (15-frame)} 
& LSTM-3Class & \textbf{0.9270} \\ \hline
\multirow{1}{*}{Ours (30-frame)} 
& LSTM-3Class & \textbf{0.9371} \\ \hline
\multirow{1}{*}{Ours (45-frame)} 
& LSTM-3Class & \textbf{0.9484} \\ \hline
\end{tabular}
\end{table*}
\setlength{\tabcolsep}{1.4pt}


\subsection{Computational Efficiency}

Execution times\footnotemark[1] \footnotetext[1]{All evaluations are done on a PC with Ubuntu 16.04, i7-7700K CPU, 16 GB RAM and an Nvidia GeForce GTX 1080 GPU.}of our LSTM models for 30-frame input sequences can be seen in Table~\ref{table:efficiency}. As shown, vehicle detection and tracking steps of our pipeline take much more time than the classification step, which is not more than 2 milliseconds. Since vehicle detection step also exists in previous works before classification step, our approach has the advantage of having just one layer and fewer parameters in classification step.

\setlength{\tabcolsep}{1.5pt}
\setlength{\extrarowheight}{.10em}
\begin{table}[h!]
\small
\begin{center}
\caption{Execution time comparison of evaluated LSTM methods}
\label{table:efficiency}
\begin{tabular}{|c|c|c|c|}
\hline
Method & \begin{tabular}[c]{@{}c@{}}Vehicle Detection \\ and Tracking \\ (sec/seq) \end{tabular} & \begin{tabular}[c]{@{}c@{}}Classification \\  (msec/seq)\end{tabular} &\begin{tabular}[c]{@{}c@{}} Total \\ (sec/seq)\end{tabular} \\ \hline
Baseline & \multirow{3}{*}{4.002} & 2.11 & 4.0041 \\
LSTM-3class & & 1.29 & 4.0032 \\
LSTM-2classLR & & 1.95 & 4.0039 \\ \hline
\end{tabular}
\end{center}
\end{table}
\setlength{\tabcolsep}{1.5pt}

Computation times are directly proportional to the number of frames. Thus, the total time is 2 seconds for 15-frame sequences and 8 seconds for 60-frame sequences. 
Please note that, in the proposed approach, we process two seconds of video regardless of the number of frames in the sequence (15, 30, 45 or 60). Vehicle detection and tracking modules can be executed as frames arrive. Thus, if we use 15-frame sequences, we are able to produce a classification result (cut-in/lane-pass) for the scene every two seconds. 
As can be seen in Tables \ref{table:comparison} and \ref{table:comparison-pr-2class}, results of 15-frame sequences are either the best or very close to the best results. From this point of view, we can argue that the proposed approach can be considered for actual implementations to detect cut-in maneuvers.

As mentioned before, we employ CNNs for vehicle detection, however different from the previous studies, we obtain features directly from the target vehicle bounding box and feed to an LSTM. This is computationally cheaper than feeding a complex CNN with video frames to extract features (\cite{Biparva2021, Llorca2020, Izquierdo2019, Yurtsever2019}). Thus, the feature extraction and classification times are longer for the methods in the literature. 


\section{Conclusion and Future Work}

The approaches developed for ADAS and autonomous vehicles should be as simple and affordable as possible. Therefore, methods that work with monocular vision, as in our study, may be preferable. In this work, we proposed a simple approach to predict possible dangerous cut-in maneuvers. Side-aware method, LSTM-2classLR, achieved a promising result (0.9585 accuracy) using just the center coordinates, width and height of the target vehicle's bounding box on BDD-100K cut-in/lane-pass subset. Furthermore, our approach exceed the current lane-change detection methods in the literature  by \%3 on accuracy metric. Last of all, when we represent last two seconds of video with 15 frames, computation time is also two seconds. Thus, we conclude that a system implementing our approach can refresh its decision every two seconds. 

As future work, we plan to increase the number of maneuver classes (e.g. a vehicle drifting in ego-lane and braking) to build more robust approach for driver assistance systems.

\section*{Acknowledgment}

This work was supported by the Scientific and Technological Research Council of Turkey (TÜBİTAK), Grant No: 2244-118C079.


{\small
\bibliographystyle{ieee_fullname}
\bibliography{egpaper}
}

\end{document}